\documentclass[10pt,twocolumn,letterpaper]{article}

\usepackage{iccv}
\usepackage{times}
\usepackage{epsfig}
\usepackage{graphicx}
\usepackage{amsmath}
\usepackage{amssymb}

\usepackage{multirow}
\usepackage{endnotes}

\usepackage[breaklinks=true,bookmarks=false]{hyperref}

\iccvfinalcopy 

\def\iccvPaperID{4009} 

\ificcvfinal\fi
\begin{document}

\title{Geometry Normalization Networks for Accurate Scene Text Detection}

\author{Youjiang~Xu\endnotemark[1]\thanks{Authors contributed equally, youjiangxu@gmail.com, duanjiaqi@hust.edu.cn}, Jiaqi~Duan\endnotemark[12]\footnotemark[1], Zhanghui~Kuang\endnotemark[1]\thanks{Corresponding author, kuangzhanghui@sensetime.com}, Xiaoyu~Yue\endnotemark[1], Hongbin~Sun\endnotemark[1], Yue~Guan\endnotemark[2], Wayne~Zhang\endnotemark[1]\\
\endnotemark[1]{SenseTime Research} \\
\endnotemark[2]{Huazhong University of Science and Technology} \\
}

\maketitle
\ificcvfinal\thispagestyle{empty}\fi

\begin{abstract}
Large geometry (\eg, orientation) variances are the key challenges in the scene text detection. In this work, we first conduct experiments to investigate the capacity of networks for learning geometry variances  on detecting scene texts, and find that networks can handle only limited text geometry variances. Then, we put forward a novel Geometry Normalization Module (GNM) with multiple branches, each of which is composed of one Scale Normalization Unit and one Orientation Normalization Unit, to normalize each text instance to one desired canonical geometry range through at least one branch.  The GNM is general and readily plugged into existing convolutional neural network based text detectors to construct end-to-end Geometry Normalization Networks (GNNets). Moreover, we propose a geometry-aware training scheme to effectively train the GNNets by sampling and augmenting text instances from a uniform geometry variance distribution. Finally, experiments on popular benchmarks of ICDAR 2015 and ICDAR 2017 MLT validate that our method  outperforms all the state-of-the-art approaches remarkably by obtaining one-forward test F-scores of 88.52 and 74.54 respectively.
\end{abstract}

\section{Introduction}

\begin{figure}
\centering
\includegraphics[width=0.93\linewidth]{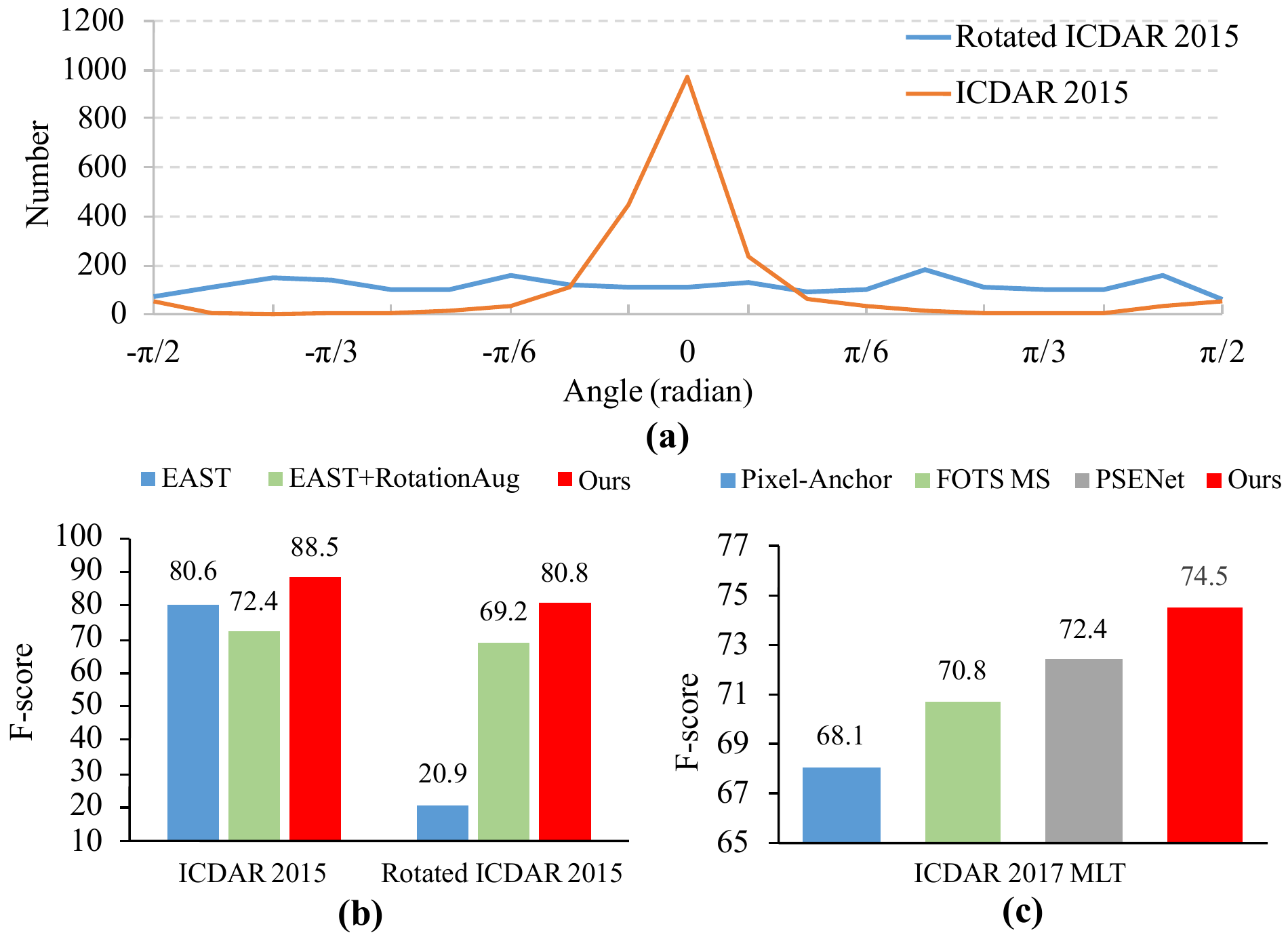}
\caption{Statistics of benchmarks and performance comparisons. (a) shows the statistics of text angles in ICDAR 2015 and the rotated ICDAR 2015; our proposed method is compared with EAST, and EAST trained with rotation augmentations on both the ICDAR 2015 and the rotated ICDAR 2015 in (b) while compared with the state-of-the-art methods on ICDAR 2017 MLT in (c). Noted that FOTS MS\cite{liu2018fots} is trained with both text bounding boxes and word recognition annotations and tested with multi-scale fusion, while ours is trained  with bounding boxes only and tested  with one single scale only. Pixel-Anchor \cite{Li2018pixel} and PSENet \cite{li2018shape} are methods proposed very recently (best viewed in color). }

\label{fig:statistics}
\end{figure}

Convolutional neural networks (CNNs) have
been dominating the research of general object detection~\cite{Girshick2014a,girshickICCV15fastrcnn,ren2015faster,Liu2016,Redmon2016}, as well as scene text detection~\cite{Tian2016,zhong2016deeptext,Liu2017,Shi2017,Zhou2017,He2017a,Liao2017,Hu2017,wang2018geometry,liu2018fots,lyu2018multi,liao2018rotation,li2018shape,Yue2018} in recent years. Thanks to the generic nature of CNN-based approaches, scene text detection can usually benefit from the rapid progress of general object detection. Despite the great success of CNNs, detecting scene texts  has its own challenges as texts have large geometry (\eg, scale or orientation) variances in real application scenarios.
Existing methods tackle the scale variance problem by detecting texts on multi-layers~\cite{Liao2017,Shi2017,He2017a} with one detection header on each or the FPN-like multi-scale fusion layer with one detection header~\cite{Zhou2017,liu2018fots,li2018shape,wang2018geometry}, and predict arbitrary
orientations by the bounding box angle estimation~\cite{liu2018fots,Zhou2017,Liao2017,He2017a} or the orientation-sensitive convolutions~\cite{wang2018geometry,liao2018rotation,He2017a}. 
Each of their individual detection header learns all training samples with enormous geometry variances or only one subset of them,
which might lead to sub-optimal performance.

\begin{figure*}[ht]
\centering
\includegraphics[width=0.95\linewidth]{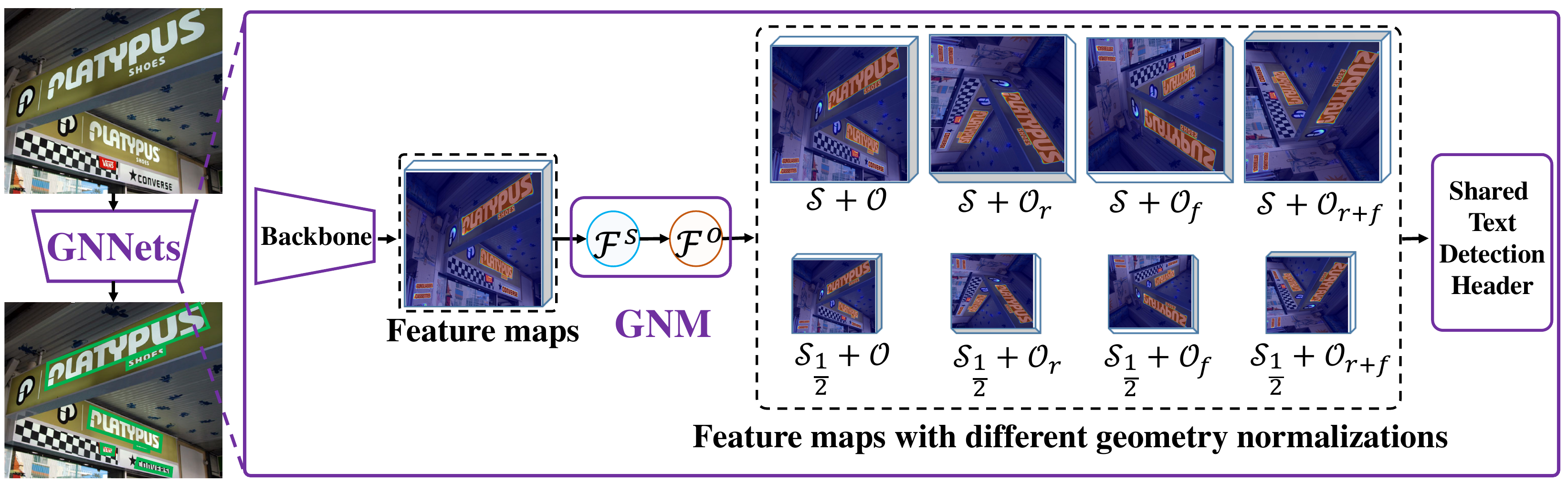}
\caption{The framework of the proposed Geometry Normalization Networks. The feature maps extracted by the backbone are fed into the Geometry Normalization Module (GNM) with  multi-branches, each of which is composed of one Scale Normalization Unit (SNU) $\mathcal{F}^s$ and Orientation Normalization Unit (ONU) $\mathcal{F}^o$. There are two different scale normalization units ($\mathcal{S}, \mathcal{S}_\frac{1}{2} $) and four orientation normalization units ($\mathcal{O}, \mathcal{O}_r, \mathcal{O}_f, \mathcal{O}_{r+f}$). With different combinations of SNU and ONU, GNM generates different geometry normalized feature maps, which are fed into one shared text detection header. }
\label{fig:framework}
\end{figure*}

Although standard multi-orientation benchmarks such as ICDAR 2015~\cite{karatzas2015icdar} and ICDAR 2017 MLT~\cite{nayef2017icdar2017} had great impact on promoting state-of-the-art text detection approaches~\cite{li2018shape,Zhou2017}, the issue of large geometry variances on text detection is overlooked by the research community.
Surprisingly,  we find that horizontal texts dominate the multi-orientation benchmark ICDAR 2015. To evaluate text detection with large geometry variances, we augment ICDAR 2015 by randomly rotating images (named by rotated ICDAR 2015\footnote{The new benchmark of rotated ICDAR 2015 will be available in: https://github.com/bigvideoresearch/GNNets}), so that the orientations are distributed uniformly for better evaluation as shown in Figure~\ref{fig:statistics} (a). As a case study, EAST~\cite{Zhou2017} degrades greatly (from 80.6\% to 20.9\%) on rotated ICDAR 2015. Even after rotation augmentation training, it still performs worse than on the original data before augmentation (see Figure~\ref{fig:statistics} (b)). We believe that this is because it cannot capture large geometry variances well.

To this end, in this paper, we conduct a series of controlled experiments to investigate the impacts of geometry variances on the scene text detection, and find that CNN based detectors can only capture limited text geometry variances but making full use of all training samples with large geometry variances could improve their generalization ability.  
To solve the above dilemma, we propose a novel Geometry Normalization Module (GNM). It has multiple normalization branches, each of which is composed of one Scale Normalization Unit (SNU) and one Orientation Normalization Unit (ONU), and can learn geometry-specific feature maps.  The geometry of each scene text instance can be transformed into one desired canonical geometry range through at least one branch of GNM. In this way, large geometry variances of all training samples are normalized to a limited distribution, so that we can train one shared text detection header on them effectively. The proposed GNM is general, and readily plugged into any CNN-based text detector to construct end-to-end Geometry Normalization Networks (GNNets). The general flowchart of GNNets is illustrated in Figure~\ref{fig:framework}. We demonstrate its superiority by equipping state-of-the-art text detectors EAST~\cite{Zhou2017} and PSENet~\cite{li2018shape} with our proposed GNM.

Training GNNets is non-trivial. We further propose a geometry-aware training strategy to effectively train GNNets by randomly augmenting text instances so that they are sampled from a uniform geometry variance distribution. In this way, all branches in the GNM have an equal number of valid text samples in each batch, and thus can be trained uniformly.

Thanks to the intrinsic geometry normalization capability of the GNM, and the proposed effective training strategy, our GNNets outperform
state-of-the-art scene text detection methods by impressive  margins.   Specifically,  on  ICDAR 2015 and ICDAR 2017 MLT,
GNNets achieve the best one-forward test F-Scores of 88.50 and 74.54 respectively (see Figure~\ref{fig:statistics} (c)), which is even better than end-to-end word spotters such as FOTS MS~\cite{liu2018fots} trained with word recognition supervision signals, and time-consuming multi-scale test methods such as~\cite{liu2018fots, Zhou2017}. Our model also outperforms methods \cite{Li2018pixel, li2018shape} that were proposed very recently.


\section{Related Work}

\textbf{Scene text detection.} Detecting scene text in the wild has received great attentions in recent years. Lots of approaches~\cite{Epshtein2010,neumann2010method, Neumann2012, Huang2013,uchida2014text,zhang2015symmetry, busta2015fastext,yin2015multi, Tian2016,zhong2016deeptext, Shi2017,Zhou2017,Hu2017,Liu2017,Jiang2017,Liao2017,liao2018rotation,lyu2018multi,liu2018fots,wang2018geometry,dai2018fused,li2018shape,Lyu2018,Yue2018} have been proposed. Comprehensive reviews can be found in the surveys~\cite{uchida2014text, ye2015text, zhu2016scene}. Herein,  we will discuss about those papers which are mostly relevant to our method in terms of geometry (scale and orientation) robustness.

 Geometry robust text detection methods target at remedying large scale variances or large orientation variances. To suppress the issue of large scale variances in text detection, inspired by SSD~\cite{Liu2016}, earlier work~\cite{Liao2017,Shi2017,He2017a} detected texts independently on multi-layers, each of which detects texts with one specific size range. Their low level features lack high level semantics, which easily leads to missing detection or false alarms. To compensate the absence of semantics in low-level features, later work~\cite{Zhou2017,liu2018fots,li2018shape,dai2018fused} introduced FPN-like~\cite{lin2017feature} or FCN-like~\cite{Long2015} architectures, and detected texts on one fused layer. The features of texts of different scales vary drastically, which hinders to learn their detection headers mapping from features to bounding boxes well.  There exist lots of multi-orientation text detection approaches~\cite{Zhou2017,Liao2017,He2017a,wang2018geometry,liao2018rotation,liu2018fots,li2018shape}. Different horizontal text detection ones,  they either directly predict the orientation of text boxes or connections between text segments~\cite{Zhou2017,Liao2017,He2017a,liao2018rotation,liu2018fots},  or construct oriented text boxes from bottom to top~\cite{li2018shape,Long2018}. To robustly detect multi-orientation texts, RRD~\cite{liao2018rotation} extracted rotation-sensitive features by explicitly rotating the convolutional filters.
 Different from all the aforementioned geometry robust text detection methods, which focus on either scale or orientation robustness in isolation, our proposed GNNets achieve both scale and orientation robustness of text detection in a unified framework.
 Recently, ITN~\cite{wang2018geometry} predicted an affine transformation for each location, which guides to deform the convolutional filters accordingly, and achieve geometry (including scale and orientation) robust text detection. However, the affine transformation prediction error might greatly degrade the final text detection performance. In contrast, our GNM can automatically normalize each text instance to one desired canonical geometry range through at least one of its branches without any explicit transformation estimation. Experimental results validate that the proposed method significantly outperforms ITN on benchmarks.

 \textbf{Scale normalization for object detection.} General object detection can also benefit from scale normalization. SNIP~\cite{singh2018analysis} proposed a scale normalization method to train the detector of objects with one desired scale range during multi-scale training. To perform multi-scale training more efficiently, SNIPER~\cite{Singh2018} selected context regions around the ground-truth instances only and sampled background regions for each scale during training. They share the similar inspiration with our work by reducing scale variances. However, SNIP and SNIPER achieve \textit{scale} normalization by image pyramid, and thus need \textit{multi-forward} test. They suffer from inevitable increasing of inference time. Different from them, our proposed method achieves both scale and orientation normalization by feature transformation, and needs \textit{one-forward} test only. Moreover, it normalizes scale and orientation in a unified framework while SNIP and SNIPER normalizes scale only.

\section{Motivations}

Although state-of-the-art methods alleviate the problem of the scale and orientation variances by novel designs~\cite{Zhou2017, liao2018rotation, wang2018geometry}, there is still great room of improvement for large geometry variations as discussed in Section 1 and illustrated in Figure~\ref{fig:statistics} (b). In this section, we will investigate the capability of text detection networks of handling large geometry variances by more comprehensive experiments.

To evaluate the impacts of the geometry variances, we construct a common evaluation benchmark with text boxes in a specific geometry range. Three different sampling strategies are designed to construct a training dataset to examine the capacity of the detector.

The most popular dataset of ICDAR 2015~\cite{karatzas2015icdar} and a state-of-the-art text detector EAST~\cite{Zhou2017} are used in our experiments.
Particularly, we select text boxes with the short side length in $[20, 40]$ pixels and the angle in $[-\frac{\pi}{12}, +\frac{\pi}{12}]$ from ICDAR 2015 test set to form the evaluation benchmark.

\textbf{Geometry Specific Sampling (GSS). }
This sampling strategy constructs text boxes within the same geometry range as evaluation by transforming text instances into the desired geometry range and makes full use of all text box instances in the original training set.

For each image, we randomly select one text instance and then transform the image, so that the chosen text instance will lie in the desired geometry range. The target parameters including the short side length and the angle are uniformly generated in the range.

\textbf{Geometry Variance Sampling (GVS).}
This sampling strategy is similar to GSS, except that the transformed text boxes are in a larger geometry range than GSS. The resulting short side length is in [0, 90] pixels and the
angle is in $[-\frac{\pi}{2}, +\frac{\pi}{2}]$.

\textbf{Limited Geometry  Specific Sampling (LGSS).}
This sampling strategy only selects a subset of the training set, which are in the same geometry range as evaluation. Compared with GSS, fewer text instances are included for training a detector.

\begin{table}[b]
\renewcommand{\arraystretch}{1.1}
\centering
\caption{Experimental results of the impacts of geometry variances. `GSS', `GVS' and `LGSS' denote `Geometry Specific Sampling', `Geometry Variance Sampling' and `Limited Geometry  Specific Sampling', respectively.}\label{table:investigation}
    \begin{tabular}{|l|c|c|c|}
      \hline

       Setting  & Recall & Precision  & F-score \\
       \hline
       GSS &  52.22 & 73.97 & 61.22 \\
       GVS & 42.80 & 57.57 & 49.10 \\
       LGSS & 42.34 & 75.44 & 54.24 \\
      \hline
      \end{tabular}
\end{table}

We summarize our observations as follows (see Table~\ref{table:investigation} for all experimental results):
\begin{itemize}
\item \textit{Existing text detectors have limited learning capacity for handling large geometry variances}. Samples with large geometry variances degrade the performance of text detectors significantly. EAST~\cite{Zhou2017} trained with GVS performs 12\% worse than GSS on F-score. This observation is also common for other state-of-the-art detectors and we do not list more results due to the space limit.
\item \textit{More transformed samples are helpful for training a geometry specific detector}. Compared with LGSS, GSS provides more training samples by transformation and brings approximately 7\% performance gain on F-score.
\end{itemize}



The above observations motivate us to normalize geometry variances during training, so that a geometry specific detector only needs to handle limited geometry range. Instead of transforming images, which suffers from time-consuming multi-scale multi-orientation feature extraction in both training and test, we design efficient geometry normalization by feature transformation. In order to fully utilize training samples, we randomly augment all text instances, so that each geometry specific detector can learn from all text instances in all images.

\section{Geometry Normalization Networks}

In this section, we will first introduce the proposed Geometry Normalization Module (GNM), and then describe overall architecture of Geometry Normalization Networks (GNNets). Finally, the geometry-aware training strategy will be presented.

\subsection{Geometry Normalization Module}

GNM targets at normalizing text geometry distribution with large variances into one desired canonical geometry range, so that text detection header can be learned well.
We divide the geometry distribution with large variance into multiple appropriate geometry distributions with small variance, each of which is allocated one independent geometry-specific branch to handle. Formally, GNM is given by
\begin{align}
        \tilde{\textbf{x}}_i = \mathcal{F}_i^o(\mathcal{F}_i^s(\textbf{x})) \quad i \in \{1,\dots,N\},  \label{eq:GNL}
\end{align}
where $\textbf{x}$ is the input feature maps of the GNM, and $\tilde{\textbf{x}}_i$ is the output feature maps of its $i^{th}$ branch.
$\mathcal{F}_i^s$ and $\mathcal{F}_i^o$  are the scale normalization unit and the orientation normalization unit of the $i^{th}$ branch, which are designed to normalize scale variances and orientation variances respectively.


\textbf{Scale Normalization Unit.}  Each scale normalization unit extracts scale-specific features. We design it by explicitly downsampling feature maps and thus expanding the
receptive field of the convolutional kernels.
Concretely, given the input feature maps $\textbf{x} \in \mathbb{R}^{C\times H\times W}$ outputted by feature extractors, the scale normalization unit outputs  ${\textbf{x}}^s\in \mathbb{R}^{C^{'}\times H^{'}\times W^{'}}$ as follows:
\begin{align}
        {\textbf{x}}^s = \mathcal{F}^s(\textbf{x}), \quad \mathcal{F}^s \in \{ \mathcal{S}, \mathcal{S}_\frac{1}{2} \}, \label{eq:SNU}
\end{align}
where $\mathcal{S}$ is one $1\times 1$ conv operator, and $\mathcal{S}_{\frac{1}{2}}$ is a stack of $1\times 1$ conv, $2\times 2$ max-pooling with stride 2, and $3\times 3$ conv operators. $\mathcal{S}$ preserves the spatial resolution of the input feature maps (\ie, $H^{'}=H$ and $W^{'}=W$), and $\mathcal{S}_\frac{1}{2}$ halves it (\ie, $H^{'}=\frac{1}{2}H$ and $W^{'}=\frac{1}{2}W$). We select two scale normalization units, because adding more units achieve marginal performance improvement as shown in experimental results.

\begin{figure}
\centering
\includegraphics[width=0.95\linewidth]{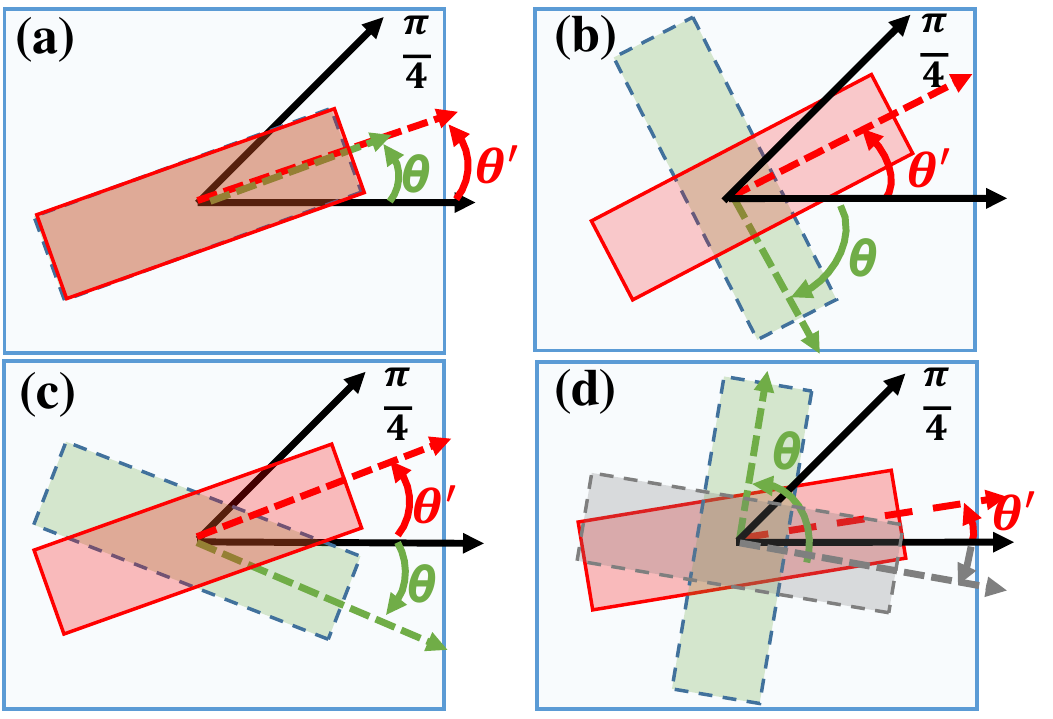}
\caption{The changes of the text box orientation by applying the ONU. The `green' box is the original box, the 'grey' box is the intermediate box during transformation and the `red' box is the result of the ONU (\eg, $\mathcal{O}, \mathcal{O}_r, \mathcal{O}_f, \mathcal{O}_{r+f}$). $\theta$ and $\theta^{'}$ are the angle of the original box and the result box, respectively. (a), (b), (c) and (d) are the procedure of $\mathcal{O}, \mathcal{O}_r, \mathcal{O}_f, \mathcal{O}_{r+f}$, respectively. And they transform texts with angles in $[0,\frac{\pi}{4}]$, $[-\frac{\pi}{2},-\frac{\pi}{4}]$, $[-\frac{\pi}{4},0]$, and $[\frac{\pi}{4},\pi]$ to those with angles in $[0,\frac{\pi}{4}]$.}
\label{fig:ONU}
\end{figure}

\textbf{Orientation Normalization Unit.} Each orientation normalization unit normalizes texts within a specific angle range to near-horizontal texts, by explicitly rotating and/or
flipping operations. Given the input feature maps ${\textbf{x}}^s\in \mathbb{R}^{C^{'}\times H^{'}\times W^{'}}$,
it outputs $\tilde{\textbf{x}} \in \mathbb{R}^{\tilde{C}\times \tilde{H}\times \tilde{W}}$ as follows:
\begin{align}
        \tilde{\textbf{x}} = \mathcal{F}^o({\textbf{x}}^s), \quad \mathcal{F}^o \in \{ \mathcal{O}, \mathcal{O}_r, \mathcal{O}_f, \mathcal{O}_{r+f}\}, \label{eq:ONM}
\end{align}
where $\mathcal{O}$ is a $1\times 1$ conv operator,  $\mathcal{O}_r$ denotes a stack of $1\times 1$ conv, rotation, and $3 \times 3$ conv operators,
$\mathcal{O}_f$ denotes a stack of $1\times 1$ conv, flipping, and $3 \times 3$ conv operators, and $\mathcal{O}_{r+f}$ denotes a sequence of $1\times 1$ conv, rotation, flipping and $3 \times 3$ conv. Rotation and flipping indicate clockwise rotating by $\frac{\pi}{2}$  and horizontal flipping, respectively. As illustrated in Figure \ref{fig:ONU}, $\mathcal{O}$, $\mathcal{O}_r$, $\mathcal{O}_f$, and $\mathcal{O}_{r+f}$ transform texts with angles in $[0,\frac{\pi}{4}]$, $[-\frac{\pi}{2},-\frac{\pi}{4}]$, $[-\frac{\pi}{4},0]$, and $[\frac{\pi}{4},\pi]$ to those with angles in $[0,\frac{\pi}{4}]$, respectively.

\begin{table*}[ht]
\centering
\footnotesize
	
	\caption{Proxy ground truth. $(x^{'}, y^{'}), h^{'}, w^{'}$, and $\theta^{'}$ denote the center, height, width and angle of the transformed text box, respectively. And $(x, y), h, w$, and $\theta$ are the center, height, width and angle of the input text box, respectively. $H$ and $W$ denote height and width of the input image.}
	\label{tab:tab_pgt}
    \begin{tabular}{c|c|c|c|c|c|c}
    \hline
	$\mathcal{F}^s$&$\mathcal{F}^o$&$x^{'}$&$y^{'}$&$h^{'}$&$w^{'}$&$\theta^{'}$ \\ \hline
    $\mathcal{S}$&$\mathcal{O}$                & $x$    &$y$ & $h$&$w$&$\theta$ \\ \hline 
    $\mathcal{S}$&$\mathcal{O}_{r}$          &$H-y$  &$x$ & $h$&$w$& $\begin{cases}\theta-\pi/2 &\theta \in [0,\pi/2],\\\theta + \pi/2 &\theta \in [-\pi/2,0).\end{cases}$ \\ \hline 
    $\mathcal{S}$&$\mathcal{O}_{f}$          &$x$      &$H-y$ & $h$&$w$&$-\theta$ \\ \hline 
    $\mathcal{S}$&$\mathcal{O}_{r+f}$       &$H-y$   &$W-x$ & $h$&$w$&$\begin{cases}-\theta+\pi/2 &\theta \in [0,\pi/2],\\-\theta - \pi/2 &\theta \in [-\pi/2,0).\end{cases}$ \\ \hline 

    $\mathcal{S}_\frac{1}{2}$&$\mathcal{O}$& $x/2$  &$y/2$ & $h/2$&$w/2$&$\theta$\\ \hline 
    $\mathcal{S}_\frac{1}{2}$&$\mathcal{O}_{r}$&$H/2-y/2$ &$x/2$ & $h/2$&$w/2$&$\begin{cases}\theta-\pi/2 &\theta \in [0,\pi/2],\\\theta + \pi/2 &\theta \in [-\pi/2,0).\end{cases}$ \\ \hline 
    $\mathcal{S}_\frac{1}{2}$&$\mathcal{O}_{f}$& $x/2$&$H/2-y/2$ & $h/2$&$w/2$&$-\theta$ \\ \hline 
    $\mathcal{S}_\frac{1}{2}$&$\mathcal{O}_{r+f}$& $H/2-y/2$&$ W/2-x/2$ & $h/2$&$w/2$&$\begin{cases}-\theta+\pi/2 &\theta \in [0,\pi/2],\\-\theta - \pi/2 &\theta \in [-\pi/2,0).\end{cases}$ \\ \hline 
	\end{tabular}
\end{table*}

\subsection{Architecture of GNNets}
The proposed GNM is general, and readily plugged into existing CNN based text detectors to construct Geometry Normalization Networks (GNNets). Figure~\ref{fig:framework} illustrates the general architecture of GNNets.
First,   images are fed into a CNN based feature extractor to obtain feature maps, which will be the input of the proposed GNM. Then, GNM  produces different scale and orientation specific feature maps, and a shared text detection header predicts text boxes on these feature maps. Finally, the predicted text boxes are transformed back accordingly and merged via NMS.

In order to demonstrate the generalization ability of our proposed method, we instantiate our GNNets with two state-of-the-art text detectors, \ie, EAST~\cite{Zhou2017} and PSENet~\cite{li2018shape}. We insert our GNM after the last layer of the feature merged branch in EAST, and the concatenation layer of the multi-scale layers of the FPN in PSENet. We optimize the following loss:
\begin{align}
        L = \frac{1}{N}\sum_{n=1}^{N}L_n(\hat x,\hat y,\hat h,\hat w,\hat \theta; x^{'},y^{'},h^{'},w^{'},\theta^{'}),   \label{eq:loss_function}
\end{align}
where $N$ denotes the number of branches.  $(\hat x,\hat y)$, $\hat h$, $\hat w$, and $\hat \theta$ indicate the center, height, width and angle of the predicted text bounding box, respectively. $(x^{'}, y^{'})$, $h^{'}$, $w^{'}$ and $\theta^{'}$ denote center, height, width and angle of the proxy ground truth, respectively, and its relation to the input bounding box $(x, y ,h, w, \theta)$ is given in Table~\ref{tab:tab_pgt}.
$L_n(\cdot)$ denotes the loss of the $n$-th geometry normalization branch of the proposed GNM, which simply follows the loss functions in EAST~\cite{Zhou2017} and PSENet~\cite{li2018shape}. Both EAST based GNNets and PSENet based GNNets outperform its counterpart significantly as shown in Section~\ref{sec:sec_exp}.


\subsection{Geometry-Aware Training and Test Strategy}\label{training_and_test_strategy}
Together with geometry normalization module, we co-design  three critical strategies so that GNNets perform well.

\textbf{Feasible geometry range.} We allocate one feasible geometry range including one scale interval and one orientation interval  to each branch of GNM.
The union of the feasible geometry ranges of all branches  equals to the whole text geometry distribution. In this way, any text bounding boxes can fall into
at least one feasible geometry range and thus can be normalized to the canonical geometry range through its corresponding branch in GMN. We will discuss the allocation in detail in Section~\ref{sec:ablation}.

\textbf{Training sampling strategy.} During training, we randomly samples one text instance, and augment it by rotating and resizing 7 times, so that each branch of the proposed GNM
has valid text instances in each batch. In this way, all branches of GNM are trained uniformly. The feature maps of all branches are used to train the text detection header of GNNets.
All the text instances in one branch  are ignored if their ground truth are not in its feasible geometry range during training.

\textbf{Test strategy.} During testing, we predict text bounding boxes on the transformed feature maps output by all branches in GNM, which are back-projected to the original scale and orientation accordingly. The output bounding boxes which do not lie in a branch's corresponding feasible geometry range  are unreliable and discarded. The remaining bounding boxes are merged via NMS.

\section{Experiments}
\label{sec:sec_exp}
In this section, we first perform ablation studies for the proposed GNNets, and then compare the GNNets with  the state-of-the-art methods. The experiments are conducted on four datasets: ICDAR 2013 \cite{karatzas2013icdar}, ICDAR 2015 \cite{karatzas2015icdar}, ICDAR 2017 MLT \cite{nayef2017icdar2017} and the Rotated ICDAR 2015.

\subsection{Benchmark Datasets}
\label{Sec:benchmark}
\textbf{ICDAR 2015} is a dataset proposed in the Challenge 4 of the 2015 Robust Reading Competition for incidental scene text detection. There are 1,000 images and 500 images for training and test, respectively. The text instances are annotated by word-level quadrangles.

\textbf{ICDAR 2017 MLT} is a dataset provided for ICDAR 2017 competition on multi-lingual scene text detection. This dataset consists of complete scene images from 9 languages. Some languages are labeled in word-level such as English, Bangla, French and Arabic, while others are labeled in line-level such as Chinese, Japanese and Korean. This dataset provides 7,200 images for training, 1,800 images for validating, and 9,000 images for testing. 

\textbf{Rotated ICDAR 2015} is constructed from the standard benchmark of ICDAR 2015. By randomly rotating images from the standard ICDAR 2015, the orientations in Rotated ICDAR 2015 are distributed uniformly (see Figure \ref{fig:statistics}). There are also 1,000 images and 500 images for training and testing, respectively. Moreover, the partitions of the Rotated ICDAR 2015 stay the same with the standard one.

\textbf{ICDAR 2013} contains 229 training images and 223 testing images. Different above dataset, this dataset only contains horizontal text instances. We  utilize the training images in the experiments of ablation studies only.

\subsection{Implementation Details}\label{Sec:Implementation}
The overall architecture of GNNets has shown in Figure \ref{fig:framework}. We use the same data augmentation techniques with previous methods \cite{Zhou2017, li2018shape}.
We further perform `Rotation Augmentation (RA)', `Initializing (Init)' and `Training sampling strategies (TSS)' to train the proposed GNNets. Particularly, `Rotation Augmentation' means that the input images are randomly rotated from $-\pi/2$ to $\pi/2$. `Initializing' means that we initialize the proposed GNNets with the reimplemented baseline models, rather than the models pre-trained on ImageNet dataset. `Training sampling strategies'  has been described in the Section~\ref{training_and_test_strategy}. We use ADAM \cite{kingma2014adam} to optimize the parameters of GNNets. For training EAST-based GNNets, the learning rate starts at 1e-4, declines to its $\frac{1}{10}$ for every 27,300 iterations, and stops at 1e-6. As for training PSENet-based GNNets, the learning rate decays per 200 epochs and there are total of 600 epochs. As for the ablation studies, we use the ICDAR 2013, the ICDAR 2015 and the Rotated ICDAR 2015 as our training data. When comparing with the state-of-the-art methods, we use the IDCAR 2015 and the ICDAR 2017 MLT as our training data, which is the same as \cite{li2018shape}.

\begin{table}[t]
\renewcommand{\arraystretch}{1.1}
\small
\centering
\caption{The impacts of Scale Normalization Unit with different desired canonical scale ranges on the ICDAR 2015. `ET' and `PT' denote `EAST' and `PSENet', respectively.}\label{table:different-scales}
    \begin{tabular}{|l|c|c|c|c|}
      \hline
       Model & Range & Recall & Precision  & F-score \\
      \hline\hline
      ET+$\mathcal{S}$ & $[10, 200]$ & 74.96 & 87.23 & 80.63 \\
      ET+$\mathcal{S}$+$\mathcal{S}_{\frac{1}{2}}$ & $[10, 100]$ & 82.42 & 87.34 & 84.81 \\
      ET+$\mathcal{S}$+$\mathcal{S}_{\frac{1}{2}}$+$\mathcal{S}_{\frac{1}{4}}$ & $[10, 60]$ & 81.12 & 85.31 & 83.16 \\

      \hline\hline
      PT+$\mathcal{S}$ & $[10, 200]$ & 83.53 & 86.10 & 85.04 \\
      PT+$\mathcal{S}$+$\mathcal{S}_{\frac{1}{2}}$ & $[10, 100]$ & 85.02 & 86.90 & 85.95 \\
      PT+$\mathcal{S}$+$\mathcal{S}_{\frac{1}{2}}$+$\mathcal{S}_{\frac{1}{4}}$ & $[10, 60]$ & 83.48 & 87.62 & 85.50 \\

      \hline
    \end{tabular}
\end{table}

\subsection{Ablation Studies}\label{sec:ablation}


\begin{table}[ht]
\renewcommand{\arraystretch}{1.1}
\centering
\footnotesize
\caption{The impacts of Orientation Normalization Unit with different desired canonical orientation ranges on the Rotated ICDAR 2015. `ET' and `PT' denote `EAST' and `PSENet', respectively.}\label{table:different-orientations}
    \begin{tabular}{|l|l|c|c|c|}
      \hline
       Model & Range & Recall & Precision  & F-score \\
      \hline\hline
      ET+$\mathcal{O}$ & $ [-\frac{\pi}{2}, \frac{\pi}{2}] $ & 63.64 & 75.89 & 69.23 \\
      ET+$\mathcal{O}$+$\mathcal{O}_r$ & $[-\frac{\pi}{4}, \frac{\pi}{4}] $ & 71.64 & 81.26 & 76.15 \\
      ET+$\mathcal{O}$+$\mathcal{O}_f$ & $[0, \frac{\pi}{2}] $  & 70.10 & 79.78 & 74.62 \\
      ET+$\mathcal{O}$+$\mathcal{O}_r$+$\mathcal{O}_f$+$\mathcal{O}_{r+f}$ & $[0, \frac{\pi}{4}] $ & 72.89 & 80.32 & 76.42 \\

      \hline\hline
      PT+$\mathcal{O}$ & $ [-\frac{\pi}{2}, \frac{\pi}{2}] $ & 73.61 & 81.63. & 77.41 \\
      PT+$\mathcal{O}$+$\mathcal{O}_r$ & $[-\frac{\pi}{4}, \frac{\pi}{4}]$ & 75.20 & 82.55 & 78.71 \\
      PT+$\mathcal{O}$+$\mathcal{O}_f$ & $[0, \frac{\pi}{2}] $ & 72.36 & 83.55 & 77.55 \\
      PT+$\mathcal{O}$+$\mathcal{O}_r$+$\mathcal{O}_f$+$\mathcal{O}_{r+f}$ & $[0, \frac{\pi}{4}] $ & 74.62 & 83.87 & 78.98 \\
      \hline
    \end{tabular}
\end{table}

\textbf{The impacts of SNU with different desired canonical scale ranges.} We study the impacts of SNU with different desired canonical scale ranges on the ICDAR 2015.
The feasible scale range of single branch SNU (\ie, $\mathcal{S}$) is set to $[10, 200]$ pixels, which means that the text boxes will be ignored if the length of their shortest size is not in the range. Note that for single branch SNU, the desired canonical scale range is the same as its feasible scale range. As for the SNU with two branches (\ie, $\mathcal{S}$+$\mathcal{S}_{\frac{1}{2}}$), we set the feasible scale range as $[10, 80]$ and $[60, 200]$ for $\mathcal{S}$ and $\mathcal{S}_{\frac{1}{2}}$, respectively. As $\mathcal{S}_{\frac{1}{2}}$ halves the feature maps, its desired canonical scale range should be $[30, 100]$. The final desired canonical scale range can be obtained by choosing the real minimum and real maximum desired canonical scales, and thus it should be $[10, 100]$. In this way, the scale range of text boxes is normalized from $[10, 200]$ (\ie, $\mathcal{S}$) to $[10, 100]$ (\ie, $\mathcal{S}$+$\mathcal{S}_{\frac{1}{2}}$) and all normalized text boxes will be used to train the same box detection header.
For SNU with three branches (\ie, $\mathcal{S}$+$\mathcal{S}_{\frac{1}{2}}$+$\mathcal{S}_{\frac{1}{4}}$), the feasible scale ranges are set as  $[10, 60]$, $[40, 100]$ and $[80, 200]$ for $\mathcal{S}$, $\mathcal{S}_{\frac{1}{2}}$ and $\mathcal{S}_{\frac{1}{4}}$, respectively. As a result, the desired canonical scale range of $\mathcal{S}$+$\mathcal{S}_{\frac{1}{2}}$+$\mathcal{S}_{\frac{1}{4}}$ is further normalized to $[10, 60]$. All results are shown in Table \ref{table:different-scales}.
Compared with three SNU branches, both `EAST+$\mathcal{S}$+$\mathcal{S}_{\frac{1}{2}}$' and `PSENet+$\mathcal{S}$+$\mathcal{S}_{\frac{1}{2}}$' achieve a better performance on the ICDAR 2015. Thus, we choose SNU with two branches as our default setting.


\begin{table*}[ht]
\small
\renewcommand{\arraystretch}{1.1}
\centering
\caption{Results of GNNets under different training strategies on the ICDAR 2015 and the Rotated ICDAR 2015. RA, The proposed GNM, Init and TSS are added into the baselines step-by-step, where `RA', `Init' and `TSS' indicate `Rotation augmentation', `Initialization' and `Training sampling strategy', respectively.}\label{table:training_strategies}
    \begin{tabular}{|c|l|c|c|c|c|c|c|}
      \hline
      \multirow{2}{*}{Backbone} &
      \multirow{2}{*}{Method} &
      \multicolumn{3}{|c|}{Rotated ICDAR 2015} & \multicolumn{3}{|c|}{ICDAR 2015} \\ \cline{3-8}
      & & Recall & Precision  & F-score & Recall & Precision  & F-score \\ \cline{1-8}

      \multirow{7}{*}{EAST}
      & (a)Reimplemented EAST (Baseline) & 12.52 & 65.00 & 20.99 & 74.96 & 87.23 & 80.63 \\ \cline{2-8}
      & (b)+RA                                                    & 63.64 & 75.89 & 69.23 & 66.28 & 79.72 & 72.43 \\
      & (c)+RA+GNM                                          & 58.01 & 79.59 & 67.11 & 63.79 & 74.57 & 68.76 \\
      & (d)+RA+GNM+Init                                   & 67.06 & 85.46 & 75.15 & 69.62 & 81.56 & 75.12 \\
      & (e)+RA+ONU+Init+TSS                          & 72.89 & 80.32 & 76.42 & 73.04 & 84.32 & 78.28 \\
      & (f)+RA+SNU+Init+TSS                           & 71.52 & 75.50 & 73.47 & 78.96 & 76.70 & 77.81 \\
      & (g)+RA+GNM+Init+TSS (ours GNNets) & 77.28 & 84.69 & 80.82 & 80.45 & 83.67 & 82.03 \\
      \hline
      \multirow{7}{*}{PSENet}
      & (a)Reimplemented PSENet (Baseline) & 56.30 & 72.04 & 63.22 & 83.53 & 86.10 & 85.04 \\ \cline{2-8}
      & (b)+RA                                                   & 73.61 & 81.63 & 77.41 & 83.28 & 83.24 & 83.26 \\
      & (c)+RA+GNM                                        & 69.47 & 82.71 & 75.29 & 78.52 & 82.54 & 80.48 \\
      & (d)+RA+GNM+Init                                 & 72.96 & 82.10 & 77.92 & 81.41 & 86.89 & 84.06 \\
      & (e)+RA+ONU+Init+TSS                        & 74.62 & 83.87 & 78.98 & 82.95 & 85.33 & 84.13 \\
      & (f)+RA+SNU+Init+TSS                          & 74.57 & 82.48 & 78.33 & 83.96 & 86.03 & 84.99 \\
      & (g)+RA+GNM+Init+TSS (ours GNNets)& 75.58 & 83.24 & 79.23 & 82.37 & 88.01 & 85.10 \\
      \hline
    \end{tabular}
\vspace{-4mm}
\end{table*}

\textbf{The impacts of ONU with different desired canonical orientation ranges.} 
We set the feasible orientation range for our proposed single branch ONU (\ie, $\mathcal{O}$) as $[-\frac{\pi}{2}, \frac{\pi}{2}]$. Note that for single branch ONU $\mathcal{O}$, the desired canonical orientation range is the same as its feasible orientation range. As $\mathcal{O}_r$ can transform text boxes with angles in $[-\frac{\pi}{2}, -\frac{\pi}{4}]$ and $[\frac{\pi}{4}, \frac{\pi}{2}]$ to those with angles in $[0, \frac{\pi}{4}]$ and $[-\frac{\pi}{4}, 0]$, respectively, its feasible orientation range is set as $[-\frac{\pi}{2}, -\frac{\pi}{4}]$ and $[\frac{\pi}{4}, \frac{\pi}{2}]$. Note that the feasible orientation range of $\mathcal{O}$, in this case, is set to $[-\frac{\pi}{4}, \frac{\pi}{4}]$. Therefore, the desired canonical orientation range of $\mathcal{O}$+$\mathcal{O}_r$ should be $[-\frac{\pi}{4}, \frac{\pi}{4}]$, because the boxes with angles are not in these ranges will be transformed by $\mathcal{O}_r$. Similarly, desired canonical orientation ranges of $\mathcal{O}$+$\mathcal{O}_f$ and $\mathcal{O}$+$\mathcal{O}_r$+$\mathcal{O}_f$+$\mathcal{O}_{r+f}$ are $[0, \frac{\pi}{2}]$ and $[0, \frac{\pi}{4}]$, respectively. The models with different desired canonical orientation ranges are trained and tested on the Rotated ICDAR 2015. All the experimental results are shown in Table \ref{table:different-orientations}. By narrowing the desired canonical orientation range from $[-\frac{\pi}{2}, \frac{\pi}{2}]$ to $[0, \frac{\pi}{4}]$, EAST and PSENet are improved by about 7.2\% and 1.5\% on the Rotated ICDAR 2015, respectively. Therefore, we set ONU consists of four branches ($\mathcal{O}$, $\mathcal{O}_r$, $\mathcal{O}_f$ and $\mathcal{O}_{r+f}$) for the later experiments.

From Table \ref{table:different-scales} and Table \ref{table:different-orientations}, we find that both narrowing the desired canonical scale range and narrowing the desired canonical orientation range benefit the model performance, which demonstrates that normalizing the geometry is beneficial to the training of the shared text detection header.


\textbf{Analysis of training strategies.} Rotation augmentation (RA), Initialization (Init), and Training sampling strategy (TSS) are utilized to promise the good performance of our proposed GNNets. 
All the results are shown in Table \ref{table:training_strategies} and we can conclude as follows:
\textbf{(1)} As shown in Table \ref{table:training_strategies} (a), EAST and PSENet, which is reimplemented by us and achieves similar performance with their original implementation \cite{Zhou2017, li2018shape}, obtain limited performances of 21.99\% and 63.22\% on the Rotated ICDAR 2015, respectively.
\textbf{(2)} When training with `RA' (see Table \ref{table:training_strategies} (b)), EAST and PSENet obtain better performances on the Rotated ICDAR 2015. However, this harms their performance on the ICDAR 2015 about 8.2\% and 1.8\%, respectively.
\textbf{(3)} Comparing with `(c)+RA+GNM', `(d)+RA+GNM+Init' obtains a performance gain on the Rotated ICDAR 2015 about 8.0\% for the EAST-based GNNets and 1.5\% for PSENet based model (see Table \ref{table:training_strategies} (c) and (d)).
\textbf{(4)} From Table \ref{table:training_strategies} (e) and (f), we find that removing either SNU or ONU will damage the performance on both ICDAR 2015 and Rotated ICDAR 2015, which demonstrates that our proposed GNNets can detect text instances with large geometry variances.
\textbf{(5)} When training with both `Init' and `TSS' (see Table \ref{table:training_strategies} (g)), our GNM obtains better performances on the Rotated ICDAR 2015 as well as on the ICDAR 2015. Particularly, compared with EAST baseline, GNNets obtain about 3 times performance gains on the Rotated ICDAR 2015 and about 1.4\% improvements on the ICDAR 2015. When comparing with PSENet baseline, our GNNets outperform about 16\% on the Rotated ICDAR 2015 and simultaneously keeps nearly the same performance on the ICDAR 2015.

\begin{table*}[t]
\renewcommand{\arraystretch}{1.1}
\footnotesize
\centering
\caption{Comparison with the state-of-the-art methods on both ICDAR 2015 and ICDAR 2017 MLT. The methods proposed in this paper are tested with only one-forward.
}\label{table:comp-state-of-the-art}
    \begin{tabular}{|l|c|c|c|c|c|c|c|}
      \hline
      \multirow{2}{*}{Model} & \multicolumn{4}{|c|}{ICDAR 2015} &

      \multicolumn{3}{|c|}{ICDAR 2017 MLT}  \\ \cline{2-8}
      &Recall & Precision & F-score & FPS & Recall & Precision  & F-score \\ \cline{1-8}

      CTPN \cite{yin2015multi} & 51.56 & 74.22 & 60.85 & 7.1 & - & - & -\\
      SegLink \cite{Shi2017} & 76.50 & 74.74 & 75.61 & - & - & - & -\\
      SSTD \cite{He2017a} & 73.86 & 80.23 & 76.91 & 7.7 & - & - & -\\
      WordSup \cite{Hu2017} & 77.03 & 79.33 & 78.16 & - & - & - & -\\
      EAST \cite{Zhou2017} & 78.33 & 83.27 & 80.72 & 6.52 & - & - & -\\
      ITN \cite{wang2018geometry} & 74.10 & 85.70 & 79.50 & - & - & - & -\\
      RRD \cite{liao2018rotation} & 79.0 & 85.6 & 82.2 & 6.5 & - & - & -\\
      FTSN \cite{dai2018fused} & 80.07 & 88.65 & 84.14 & - & - & - & -\\

      TDN SJTU2017 \cite{mlt} & - & - & - & - & 47.13 & 64.27 & 54.38 \\
      SARI FDU RRPN v1 \cite{mlt} & - & - & - & - & 55.50 & 71.17 & 62.37 \\
      SCUT DLVClab1 \cite{mlt} & - & - & - & - & 54.54 & 80.28 & 64.96 \\
      EAST++ \cite{mlt} & - & - & - & -  & 80.42 & 66.61 &72.86 \\

      FOTS \cite{liu2018fots} & 85.17 & \color{red} 91.00 & 87.99 & 7.5 & 57.51 & \color{red}80.95 & 67.25 \\
      Pixel-Anchor \cite{Li2018pixel} & 87.50 & 88.32 & 87.68 & 10 &59.54 & 79.54 & 68.10\\
      PSENet \cite{li2018shape} & 85.22 & 89.30 & 87.21 & 2.33 & 68.35 & 76.97 & 72.40 \\

      \hline
      PSENet (reimplemented) & 86.58 & 88.02 & 87.30 & 2.4 & 69.75 & 75.99 & 72.74 \\
      Ours GNNets& \color{red}86.71 & \color{blue} 90.41 & \color{red}88.52 & 2.1 & \color{red}70.06 & \color{blue}79.63 & \color{red}74.54 \\
      \hline
    \end{tabular}
\vspace{-3mm}
\end{table*}


\begin{figure*}[t]
\centering
\includegraphics[width=0.93\linewidth]{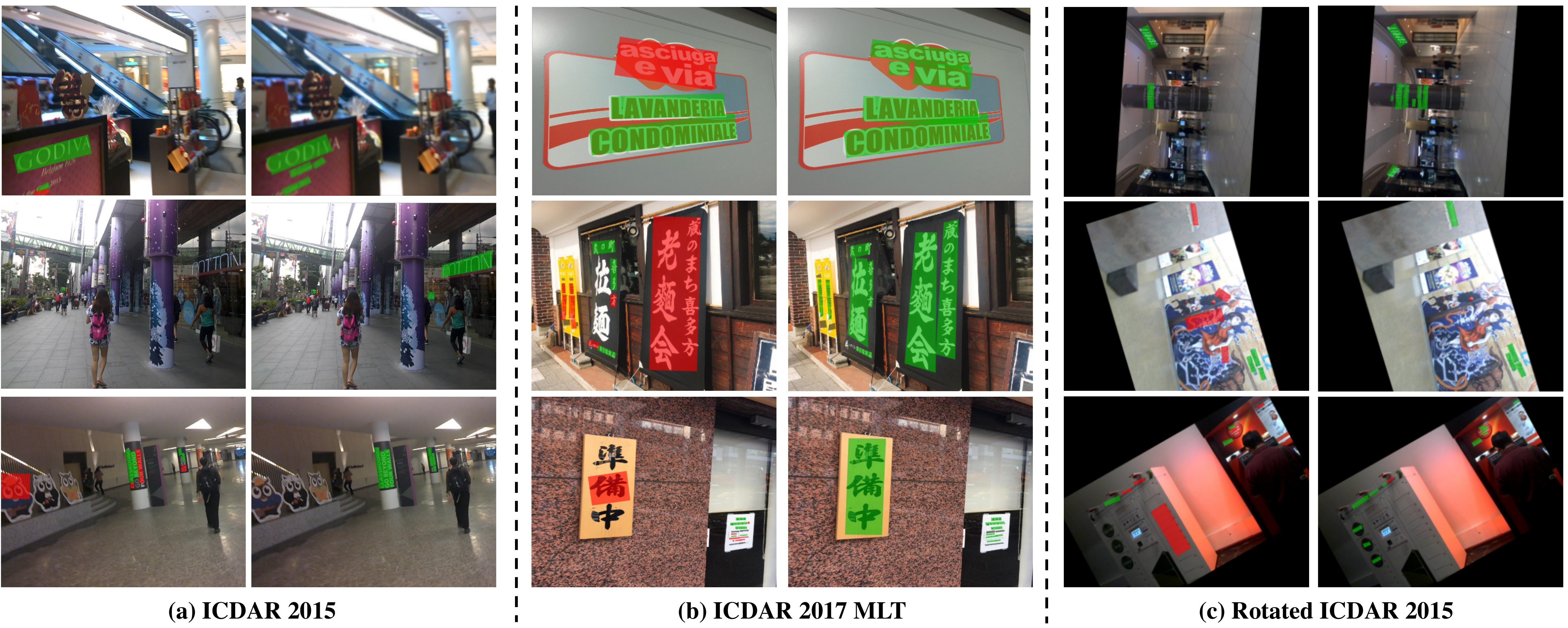}
\caption{Qualitative results on ICDAR 2015, ICDAR 2017 MLT and Rotated ICDAR 2015. We compare the results generated by PSENet (the left column) and our proposed GNNets (the right column) for each test image.}
\vspace{-3mm}
\label{fig:examples}
\end{figure*}



\subsection{Comparisons with State-of-the-Art Methods}
\label{Sec:Comparisons}

In this section, we compare our PSENet-based GNNets with several state-of-the-art methods. The same with \cite{li2018shape}, during the testing, the longer side of input images is resized to 2240 and 3200 on ICDAR 2015 and ICDAR 2017 MLT, respectively. As a large proportion of the text instances in both ICDAR 2015 and ICDAR 2017 MLT is near horizontal, the GNM contains only SNU branches. For fair comparisons, we perform only one-forward testing on these two datasets. All the results are shown in Table \ref{table:comp-state-of-the-art}.

From Table \ref{table:comp-state-of-the-art}, we can find that our reimplemented PSENet has similar performance (the differences less than 0.4\% on F-score) to the original PSENet \cite{li2018shape}. As the original PSENet \cite{li2018shape} is a very recently proposed method and its source code is still unavailable, we conduct the following experiments based on our implementation. Compared to the original PSENet \cite{li2018shape}, our GNNets achieve a performance improvement about 1.3\% and 2.1\% on ICDAR 2015 and ICDAR 2017 MLT, respectively. Compared with EAST \cite{Zhou2017} and ITN \cite{wang2018geometry} on the ICDAR 2015, our GNNets outperform them by absolute about 8\% and 9\%, respectively. While comparing with FTSN \cite{dai2018fused}, we could obtain a performance gain of 4.5\%. Our method outperforms FOTS \cite{liu2018fots} by 0.6\% on the ICDAR 2015 and by 7.3\% on the ICDAR 2017 MLT. Noted that FOTS uses text recognition annotations to benefit the network training, while we  utilize the detection annotations only.
Comparing with Pixel-Anchor \cite{Li2018pixel}, we also obtain a performance gain of 0.9\% and 6.4\% on the ICDAR 2015 and ICDAR 2017 MLT, respectively.
Finally, our GNNets with one-forward test achieves the state-of-the-art performance on both ICDAR 2015 and ICDAR 2017 MLT.

\subsection{Qualitative Results}\label{Sec:Examples}
Figure \ref{fig:examples} compares the detection results by the PSENet and our proposed GNNets on  ICDAR 2015, ICDAR 2017 MLT and Rotated ICDAR 2015. We observe that our proposed GNNets are able to detect scene text instances with large geometry variances.

\section{Conclusion}
In this paper, we put forward a novel Geometry Normalization Module (GNM) to generate several geometry aware feature maps. The proposed GNM is general and can be readily plugged into any CNN based detector to construct end-to-end Geometry Normalization Networks (GNNets). Extensive experiments illustrate that our proposed GNNets achieve an excellent performance on the detection of text instances with large geometry variances (\eg, the Rotated ICDAR 2015), and outperform the baselines with a large margin. Furthermore, our GNNets obtain a significant performance gain over the state-of-the-art methods on two popular benchmarks of ICDAR 2015 and ICDAR 2017 MLT.
\\
\textbf{Acknowledgement} This work was supported in part by Beijing Municipal Science and Technology Commission (Z181100008918004).


{\small
\bibliographystyle{ieee_fullname}
\bibliography{egbib}

\begin{thebibliography}{10}\itemsep=-1pt

\bibitem{mlt}
{ICDAR2017 Competition on Multi-Lingual Scene Text Detection and Script
  Identification}.
\newblock \url{http://rrc.cvc.uab.es/?ch=8&com=introduction}, 2017.

\bibitem{busta2015fastext}
Michal Busta, Lukas Neumann, and Jiri Matas.
\newblock {Fastext: Efficient Unconstrained Scene Text Detector}.
\newblock In {\em ICCV}, 2015.

\bibitem{dai2018fused}
Yuchen Dai, Zheng Huang, Yuting Gao, Youxuan Xu, Kai Chen, Jie Guo, and Weidong
  Qiu.
\newblock {Fused Text Segmentation Networks for Multi-Oriented Scene Text
  Detection}.
\newblock In {\em ICPR}, 2018.

\bibitem{Epshtein2010}
Boris Epshtein, Eyal Ofek, and Yonatan Wexler.
\newblock {Detecting Text in Natural Scenes with Stroke Width Transform}.
\newblock In {\em CVPR}, 2010.

\bibitem{girshickICCV15fastrcnn}
Ross Girshick.
\newblock {Fast R-CNN}.
\newblock In {\em ICCV}, 2015.

\bibitem{Girshick2014a}
Ross Girshick, Jeff Donahue, Trevor Darrell, and Jitendra Malik.
\newblock {Rich Feature Hierarchies for Accurate Object Detection and Semantic
  Segmentation}.
\newblock In {\em CVPR}, 2014.

\bibitem{He2017a}
Pan He, Weilin Huang, Tong He, Qile Zhu, Yu Qiao, and Xiaolin Li.
\newblock {Single Shot Text Detector with Regional Attention}.
\newblock In {\em ICCV}, 2017.

\bibitem{Hu2017}
Han Hu, Chengquan Zhang, Yuxuan Luo, Yuzhuo Wang, Junyu Han, and Errui Ding.
\newblock {WordSup: Exploiting Word Annotations for Character Based Text
  Detection}.
\newblock In {\em ICCV}, 2017.

\bibitem{Huang2013}
Weilin Huang, Zhe Lin, Jianchao Yang, and Jue Wang.
\newblock {Text Localization in Natural Images using Stroke Feature Transform
  and Text Covariance Descriptors}.
\newblock In {\em ICCV}, 2013.

\bibitem{Jiang2017}
Yingying Jiang, Xiangyu Zhu, Xiaobing Wang, Shuli Yang, Wei Li, Hua Wang, Pei
  Fu, and Zhenbo Luo.
\newblock R2cnn: Rotational region cnn for orientation robust scene text
  detection.
\newblock {\em arXiv preprint arXiv:1706.09579}, 2017.

\bibitem{karatzas2015icdar}
Dimosthenis Karatzas, Lluis Gomez-Bigorda, Anguelos Nicolaou, Suman Ghosh,
  Andrew Bagdanov, Masakazu Iwamura, Jiri Matas, Lukas Neumann,
  Vijay~Ramaseshan Chandrasekhar, Shijian Lu, and Others.
\newblock {ICDAR 2015 Competition on Robust Reading}.
\newblock In {\em ICDAR}, 2015.

\bibitem{karatzas2013icdar}
Dimosthenis Karatzas, Faisal Shafait, Seiichi Uchida, Masakazu Iwamura,
  Lluis~Gomez i Bigorda, Sergi~Robles Mestre, Joan Mas, David~Fernandez Mota,
  Jon~Almazan Almazan, and Lluis~Pere De~Las~Heras.
\newblock Icdar 2013 robust reading competition.
\newblock In {\em ICDAR}, 2013.

\bibitem{kingma2014adam}
Diederik~P Kingma and Jimmy Ba.
\newblock Adam: A method for stochastic optimization.
\newblock {\em arXiv preprint arXiv:1412.6980}, 2014.

\bibitem{li2018shape}
Xiang Li, Wenhai Wang, Wenbo Hou, Ruo-Ze Liu, Tong Lu, and Jian Yang.
\newblock {Shape Robust Text Detection with Progressive Scale Expansion
  Network}.
\newblock {\em arXiv preprint arXiv:1806.02559}, 2018.

\bibitem{Li2018pixel}
Yuan Li, Yuanjie Yu, Zefeng Li, Yangkun Lin, Meifang Xu, Jiwei Li, and Xi Zhou.
\newblock Pixel-anchor: A fast oriented scene text detector with combined
  networks.
\newblock {\em arXiv preprint arXiv:1811.07432}, 2018.

\bibitem{Liao2017}
Minghui Liao, Baoguang Shi, Xiang Bai, Xinggang Wang, and Wenyu Liu.
\newblock {TextBoxes: a Fast Text Detector with a Single Deep Neural Network}.
\newblock In {\em AAAI}, 2017.

\bibitem{liao2018rotation}
Minghui Liao, Zhen Zhu, Baoguang Shi, Gui-song Xia, and Xiang Bai.
\newblock {Rotation-Sensitive Regression for Oriented Scene Text Detection}.
\newblock In {\em CVPR}, 2018.

\bibitem{lin2017feature}
Tsung-Yi Lin, Piotr Doll{\'{a}}r, Ross Girshick, Kaiming He, Bharath Hariharan,
  and Serge Belongie.
\newblock {Feature Pyramid Networks for Object Detection}.
\newblock In {\em CVPR}, 2017.

\bibitem{Liu2016}
Wei Liu, Dragomir Anguelov, Dumitru Erhan, Christian Szegedy, Scott Reed,
  Cheng~Yang Fu, and Alexander~C. Berg.
\newblock {SSD: Single Shot Multibox Detector}.
\newblock In {\em ECCV}, 2016.

\bibitem{liu2018fots}
Xuebo Liu, Ding Liang, Shi Yan, Dagui Chen, Yu Qiao, and Junjie Yan.
\newblock {Fots: Fast Oriented Text Spotting with a Unified Network}.
\newblock In {\em CVPR}, 2018.

\bibitem{Liu2017}
Yuliang Liu and Lianwen Jin.
\newblock {Deep Matching Prior Network: Toward Tighter Multi-oriented Text
  Detection}.
\newblock In {\em CVPR}, 2017.

\bibitem{Long2015}
Jonathan Long, Evan Shelhamer, and Trevor Darrell.
\newblock {Fully Convolutional Networks for Semantic Segmentation}.
\newblock In {\em CVPR}, 2015.

\bibitem{Long2018}
Shangbang Long, Jiaqiang Ruan, Wenjie Zhang, and Xin He.
\newblock {TextSnake : A Flexible Representation for Arbitrary Shapes}.
\newblock In {\em ECCV}, 2018.

\bibitem{Lyu2018}
Pengyuan Lyu, Minghui Liao, Cong Yao, Wenhao Wu, and Xiang Bai.
\newblock {Mask TextSpotter: An End-to-End Trainable Neural Network for
  Spotting Text with Arbitrary Shapes}.
\newblock In {\em ECCV}, 2018.

\bibitem{lyu2018multi}
Pengyuan Lyu, Cong Yao, Wenhao Wu, Shuicheng Yan, and Xiang Bai.
\newblock {Multi-Oriented Scene Text Detection via Corner Localization and
  Region Segmentation}.
\newblock In {\em CVPR}, 2018.

\bibitem{nayef2017icdar2017}
Nibal Nayef, Fei Yin, Imen Bizid, Hyunsoo Choi, Yuan Feng, Dimosthenis
  Karatzas, Zhenbo Luo, Umapada Pal, Christophe Rigaud, Joseph Chazalon, and
  Others.
\newblock {ICDAR2017 Robust Reading Challenge on Multi-Lingual Scene Text
  Detection and Script Identification-RRC-MLT}.
\newblock In {\em ICDAR}, 2017.

\bibitem{neumann2010method}
Lukas Neumann and Jiri Matas.
\newblock {A Method for Text Localization and Recognition in Real-World
  Images}.
\newblock In {\em ACCV}, 2010.

\bibitem{Neumann2012}
Lukas Neumann and Jiri Matas.
\newblock {Real-time Scene Text Localization and Recognition}.
\newblock In {\em CVPR}, 2012.

\bibitem{Redmon2016}
Joseph Redmon, Santosh Divvala, Ross Girshick, and Ali Farhadi.
\newblock {You Only Look Once: Unified, Real-Time Object Detection}.
\newblock In {\em CVPR}, 2016.

\bibitem{ren2015faster}
Shaoqing Ren, Kaiming He, Ross Girshick, and Jian Sun.
\newblock {Faster RCNN: Towards Real-time Object Detection with Region Proposal
  Networks}.
\newblock In {\em NIPS}, 2015.

\bibitem{Shi2017}
Baoguang Shi, Xiang Bai, and Serge Belongie.
\newblock {Detecting Oriented Text in Natural Images by Linking Segments}.
\newblock In {\em CVPR}, 2017.

\bibitem{singh2018analysis}
Bharat Singh and Larry~S Davis.
\newblock {An Analysis of Scale Invariance in Object Detection SNIP}.
\newblock In {\em CVPR}, 2018.

\bibitem{Singh2018}
Bharat Singh, Mahyar Najibi, and Larry~S Davis.
\newblock {SNIPER: Efficient multi-scale training}.
\newblock In {\em NIPS}, 2018.

\bibitem{Tian2016}
Zhi Tian, Weilin Huang, Tong He, Pan He, and Yu Qiao.
\newblock {Detecting Text in Natural Image with Connectionist Text Proposal
  Network}.
\newblock In {\em ECCV}, 2016.

\bibitem{uchida2014text}
Seiichi Uchida.
\newblock {Text Localization and Recognition in Images and Video}.
\newblock {\em Handbook of Document Image Processing and Recognition}, 2014.

\bibitem{wang2018geometry}
Fangfang Wang, Liming Zhao, Xi Li, Xinchao Wang, and Dacheng Tao.
\newblock {Geometry-Aware Scene Text Detection with Instance Transformation
  Network}.
\newblock In {\em CVPR}, 2018.

\bibitem{ye2015text}
Qixiang Ye and David Doermann.
\newblock {Text Detection and Recognition in Imagery: A Survey}.
\newblock {\em PAMI}, 2015.

\bibitem{yin2015multi}
Xu-Cheng Yin, Wei-Yi Pei, Jun Zhang, and Hong-Wei Hao.
\newblock {Multi-orientation Scene Text Detection with Adaptive Clustering}.
\newblock {\em PAMI}, 2015.

\bibitem{Yue2018}
Xiaoyu Yue, Zhanghui Kuang, Zhaoyang Zhang, Zhenfang Chen, Pan He, Yu Qiao, and
  Wayne Zhang.
\newblock {Boosting up Scene Text Detectors with Guided CNN}.
\newblock In {\em BMVC}, 2018.

\bibitem{zhang2015symmetry}
Zheng Zhang, Wei Shen, Cong Yao, and Xiang Bai.
\newblock {Symmetry-based Text Line Detection in Natural Scenes}.
\newblock In {\em CVPR}, 2015.

\bibitem{zhong2016deeptext}
Zhuoyao Zhong, Lianwen Jin, Shuye Zhang, and Ziyong Feng.
\newblock {Deeptext: A Unified Framework for Text Proposal Generation and Text
  Detection in Natural Images}.
\newblock {\em arXiv preprint arXiv:1605.07314}, 2016.

\bibitem{Zhou2017}
Xinyu Zhou, Cong Yao, He Wen, Yuzhi Wang, Shuchang Zhou, Weiran He, and Jiajun
  Liang.
\newblock {EAST: An Efficient and Accurate Scene Text Detector}.
\newblock In {\em CVPR}, 2017.

\bibitem{zhu2016scene}
Yingying Zhu, Cong Yao, and Xiang Bai.
\newblock {Scene Text Detection and Recognition: Recent Advances and Future
  Trends}.
\newblock {\em Frontiers of Computer Science}, 2016.

\end{thebibliography}
}

\end{document}